\documentclass[final]{cvpr}

\usepackage{float}
\usepackage{booktabs} 
\usepackage{times}
\usepackage{epsfig}
\usepackage{graphicx}
\usepackage{amsmath}
\usepackage{amssymb}
\usepackage{array,url,kantlipsum}
\usepackage{lipsum}
\usepackage{mathtools}
\usepackage{cuted}

\usepackage{caption}

\usepackage[table,xcdraw]{xcolor}
\usepackage{tabularx}
\usepackage{enumitem}

\newcommand{\PAR}[1]{\vskip4pt \noindent {\bf #1}}
\usepackage[pagebackref=true,breaklinks=true,colorlinks,bookmarks=false]{hyperref}
\usepackage{url}            

\def\cvprPaperID{27} 
\def\confYear{CVPR 2021}
\begin{document}

\title{EarthNet2021: A large-scale dataset and challenge for Earth surface forecasting as a guided video prediction task.}

\author{Christian Requena-Mesa$^{1,2,3,}$\footnotemark[1] , Vitus Benson$^{1,}$\thanks{Joint first authors.} , Markus Reichstein$^{1,4}$, Jakob Runge$^{2,5}$, Joachim Denzler$^{2,3,4}$
\vspace{2mm}
\and
1) Biogeochemical Integration, Max-Planck-Institute for Biogeochemistry, Jena, Germany\\
2) Institute of Data Science, German Aerospace Center (DLR), Jena, Germany\\
3) Computer Vision Group, University of Jena, Jena, Germany\\
4) Michael-Stifel-Center Jena for Data-driven and Simulation Science, Jena, Germany\\
5) Technische Universität Berlin, Berlin, Germany\\
}



\twocolumn[{%
\renewcommand\twocolumn[1][]{#1}

\maketitle 
\vspace{-34pt}

\begin{center}
    \centering
    \includegraphics[width=\textwidth]{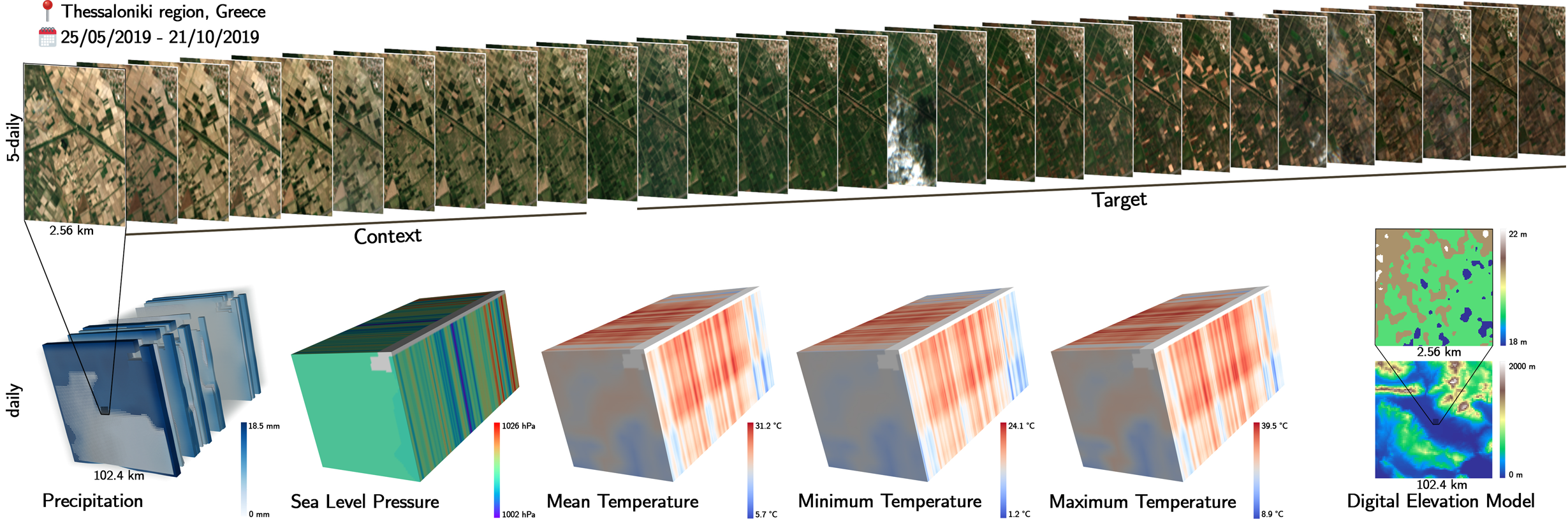}
    \vspace{-16pt}
    \captionsetup{type=figure}
    \captionof{figure}{Overview visualization of one of the over $32000$ samples in EarthNet2021.}
\end{center}%


}]

{
\renewcommand{\thefootnote}{\fnsymbol{footnote}}
\footnotetext[1]{Joint first authors.  \{crequ,vbenson\}@bgc-jena.mpg.de}}


\begin{abstract}
Satellite images are snapshots of the Earth surface. We propose to forecast them. We frame Earth surface forecasting as the task of predicting satellite imagery conditioned on future weather. EarthNet2021 is a large dataset suitable for training deep neural networks on the task. It contains Sentinel~2 satellite imagery at $20$~m resolution, matching topography and mesoscale ($1.28$~km) meteorological variables packaged into $32000$ samples. Additionally we frame EarthNet2021 as a challenge allowing for model intercomparison. Resulting forecasts will greatly improve ($>\times50$) over the spatial resolution found in numerical models. This allows localized impacts from extreme weather to be predicted, thus supporting downstream applications such as crop yield prediction, forest health assessments or biodiversity monitoring. Find data, code, and how to participate at \url{www.earthnet.tech}. 

\end{abstract}

\section{Introduction}
Seasonal weather forecasts are potentially very valuable in support of sustainable development goals such as zero hunger or life on land. 
Spatio-temporal deep learning is expected to improve the predictive ability of seasonal weather forecasting \cite{reichstein2019}. Yet it is unclear, how exactly this expectation will materialize.
One possible way can be found by carefully thinking about the target variable.
The above mentioned goals illustrate that ultimately it will not directly be the seasonal meteorological forecasts but rather derived impacts (e.g. agricultural output and ecosystem health) that are of most use to humanity. Such impacts, especially those affecting vegetation, materialize on the land surface. Meaning, they can be observed on satellite imagery.
Thus, high-resolution impact forecasting can be phrased as the prediction of satellite imagery \cite{das2016, gao2006, hong2017, lee2019, zhu2015}. Prediction of future frames is also the metier of video prediction \cite{babaeizadeh2017stochastic,finn2016unsupervised,lee2018stochastic,mathieu2015deep,oh2015action}. Yet, satellite image forecasting can also leverage additional future drivers, such as the output of numerical weather simulations with earth system models. The general setting of video prediction with additional drivers is called guided video prediction. We define \emph{Earth surface forecasting} as the prediction of satellite imagery conditioned on future weather.


Our main \textbf{contributions} are summarized as follows:
{
\begin{itemize}[noitemsep,topsep=0pt,partopsep=0pt]
    \item We motivate the novel task of Earth surface forecasting as guided video prediction and define an evaluation pipeline based the EarthNetScore ranking criterion.
    \item We introduce EarthNet2021, a carefully curated large-scale dataset for Earth surface forecasting conditional on meteorological projections.
    \item We start model intercomparison in Earth surface forecasting with a pilot study encouraging further research in deep learning video prediction models.
\end{itemize}
}

\begin{figure}[]
    \centering
    \includegraphics[width=\columnwidth]{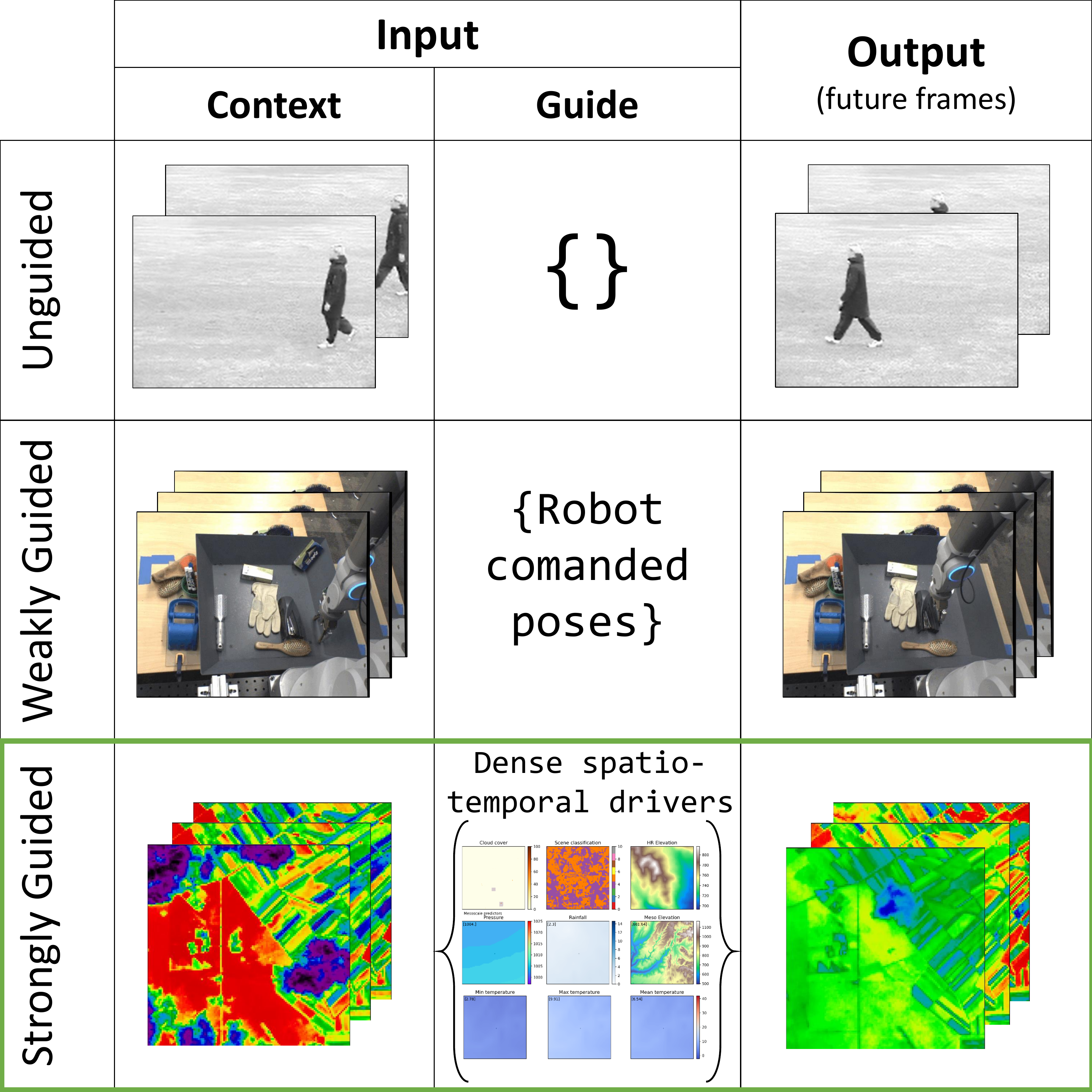}
    \caption{Video prediction could be unguided, weakly guided or strongly guided. EarthNet2021 is the first dataset specifically designed for the development of spatio-temporal strongly guided video prediction methods.}
    \label{fig:videopred}
\end{figure}

\section{Related work}
Earth surface forecasting lays in the intersection of video prediction and data-driven Earth system modeling. 

\PAR{Video prediction.}
In \figref{fig:videopred} we classify video prediction tasks into three types depending on the input data used. Traditionally video prediction models are conditioned on past context frames and predict future target frames (i.e. unguided, \cite{denton2017unsupervised,1710.05268,finn2016unsupervised,Gao_2019_ICCV,kalchbrenner2017video,kreil2020,lotter2016deep,mathieu2015deep,srivastava2015unsupervised}). 
The used models inherit many characteristics known the be useful in modeling Earth surface phenomena: short-long term memory effects \cite{kraft2019,russwurm2017temporal}, short-long range spatial relationships \cite{requena2018}, as well as, the ability to generate stochastic predictions \cite{babaeizadeh2017stochastic,franceschi2020stochastic,lee2018stochastic}, ideal to generate ensemble forecasts of Earth surface for effective uncertainty management \cite{zhu2005ensemble}.
Guided video prediction is the setting where on top of past frames models have access to future information. We further differ between weak and strong guiding (\figref{fig:videopred}). Weakly guided models are provided with sparse information of the future, for example robot commands \cite{finn2016unsupervised}. In contrast strongly guided models leverage dense spatio-temporal information of the future. This is the setting of EarthNet2021. Some past works resemble the strongly guided setting, however either the future information are derived from the frames themselves, making the approaches not suitable for prediction \cite{wang2018video} or they use the dense spatial information but discard the temporal component \cite{lutjens2020physics, requena2018}.

\PAR{Modelling weather impact with machine learning.}
Both impact modeling and weather forecasting have been tackled with machine learning methods of different complexity.
One string of literature has focused on forecasting imagery from weather satellites \cite{hong2017,lee2019,thong2017,xu2019} while another one has focused on predicting reanalysis data or emulating general circulation models \cite{rasp2020weatherbench,rasp2021data,scher2019weather,weyn2020improving}. For localized weather, statistical downscaling has been leveraged \cite{bedia2020statistical,medina2018deep,vandal2017,vandal2018} (Fig.~\ref{fig:approaches}A). Direct impacts of extreme weather have been predicted one at a time (Fig.~\ref{fig:approaches}B), examples being crop yield \cite{al2016,cai2019,kamir2020,peng2018,schwalbert2020}, vegetation index \cite{das2016,gobbi2019,ploton2017,wolanin2019}, drought index \cite{park2019} and soil moisture \cite{efremova2019}.

\section{Motivation}

While satellite imagery prediction is an interesting task for video prediction modelers, it is similarly important for domain experts, i.e., climate and land surface scientists. We focus on predicting localized impacts of extreme weather. This is highly relevant since extreme weather impacts very heterogeneously at the local scale \cite{kogan1990remote}. Very local factors, such as vegetation, soil type, terrain elevation or slope, determine whether a plot is resilient to a heatwave or not. For example, ecosystems next to a river might survive droughts more easily than those on south-facing slopes. However, the list of all possible spatio-temporal interactions is far from being mechanistically known; hence, it is a source of considerable amount of uncertainty and an opportunity for powerful data-driven methods. 

\begin{figure}[t]
    \centering
    \includegraphics[width = \columnwidth]{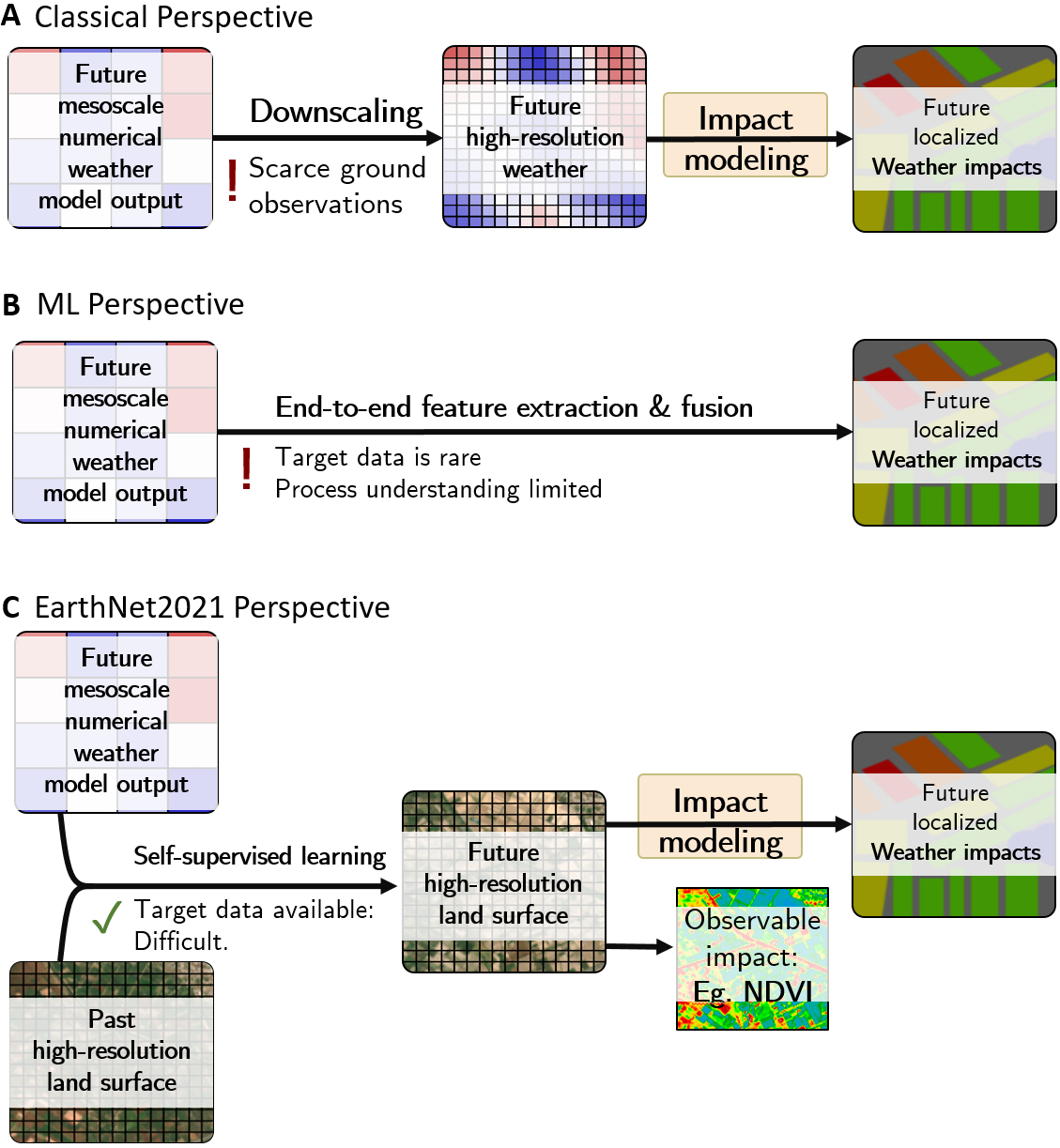}
    \caption{Three ways of extreme weather impact prediction are A) downscaling meteorological forecasts and subsequent impact modeling (\eg runoff models), B) acquiring target data at high-resolution and using supervised learning or C) leveraging Earth surface forecasting: This gives directly obtainable impacts (\eg NDVI) while still allowing for impact modeling. Compared to A) and B), large amounts of satellite imagery are available, thus together with self-supervised (i.e. no labeling required) deep learning large-scale impact prediction becomes feasible.}
    \label{fig:approaches}
\end{figure}

Predicting localized weather impacts can be tackled in three main ways (\figref{fig:approaches}). All approaches make use of seasonal weather forecasts \cite{brunet2010collaboration,cantelaube2005seasonal}($2$ -- $6$ months ahead). The classical approach (\figref{fig:approaches}A), attempts the hyper-resolution of the weather forecast for particular geolocations using statistical \cite{boe2006simple, vrac2007statistical} or dynamical \cite{lo2008assessment} downscaling, that is, correlating the past observed weather with past mesoscale model outputs and using the estimated relationship. The downscaled weather can then be used in mechanistic models (e.g. of river discharge) for impact extraction. However, weather downscaling is a difficult task because it requires ground observations from weather stations, which are sparse. A more direct way (\figref{fig:approaches}B) is to correlate a desired future impact variable, such as crop yields or flood risk, with past data (e.g., weather data and vegetation status, \cite{peng2018}). Yet again, this approach requires ground truth data of target variables, which is scarce, thus limiting the global applicability of the approach. 

Instead, by defining the task of Earth surface forecasting we propose to use satellite imagery as an intermediate step (\figref{fig:approaches}C). From satellite imagery, multiple indices describing the vegetation state such as the normalized differenced vegetation index (NDVI) or the enhanced vegetation index (EVI) can be directly observed. These give insights on vegetation anomalies, which in turn describe the ecosystem impact at a very local scale. Because of satellite imagery's vast availability, there is no data scarcity. While technically difficult, forecasting weather impacts via satellite imagery prediction is feasible. Additionally, satellite imagery is also used to extract further processed weather impact data products, such as biodiversity state \cite{fauvel2020}, crop yield \cite{schwalbert2020}, soil moisture \cite{efremova2019} or ground biomass \cite{ploton2017}. In short, Earth surface prediction is promising for forecasting highly localized climatic impacts.

\section{EarthNet2021 Dataset}
\subsection{Overview}
\PAR{Data sources.}
With EarthNet2021 we aim at creating the first dataset for the novel task of Earth surface forecasting. The task requires satellite imagery time series at high temporal and spatial resolution and additional climatic predictors. The two primary public satellite missions for high-resolution optical imagery are Landsat and Sentinel 2. While the former only revisits each location on Earth every 16 days, the latter does so every five days. Thus we choose Sentinel 2 imagery \cite{louis2016} for EarthNet2021. The additional climatic predictors should ideally come from a seasonal weather model. Obtaining predictions from a global seasonal weather model starting at multiple past time steps is computationally very demanding. Instead, we approximate the forecasts using the E-OBS \cite{cornes2018} observational dataset, which essentially contains interpolated ground truth observed weather from a number of stations over Europe. This also makes the task easier as uncertain weather forecasts are replaced with certain observations. Since E-OBS limits the spatial extent to Europe, we use the appropriate high-resolution topography: EU-DEM \cite{bashfield2011}. 

\PAR{Individual samples.}
After data processing, EarthNet2021 contains over $32000$ samples, which we call \emph{minicubes}. A single minicube is visualized in Fig.~1. It contains $30$ 5-daily frames ($128 \times 128$ pixel or $2.56 \times 2.56$ km) of four channels (blue, green, red, near-infrared) of satellite imagery with binary quality masks at high-resolution ($20$ m), $150$ daily frames ($80 \times 80$ pixel or $102.4 \times 102.4$ km) of five meteorological variables (precipitation, sea level pressure, mean, minimum and maximum temperature) at mesoscale resolution ($1.28$ km) and the static digital elevation model at both high- and mesoscale resolution.
The minicubes reveal a strong difference between EarthNet2021 and classic video prediction datasets. In the latter, the objects move in a 3d space, but images are just a 2d projection of this space. For satellite imagery, this effect almost vanishes as the Earth surface locally is very similar to a 2d space.

\begin{center}
	\begin{figure}[t]
		\centering
		\includegraphics[width=\columnwidth]{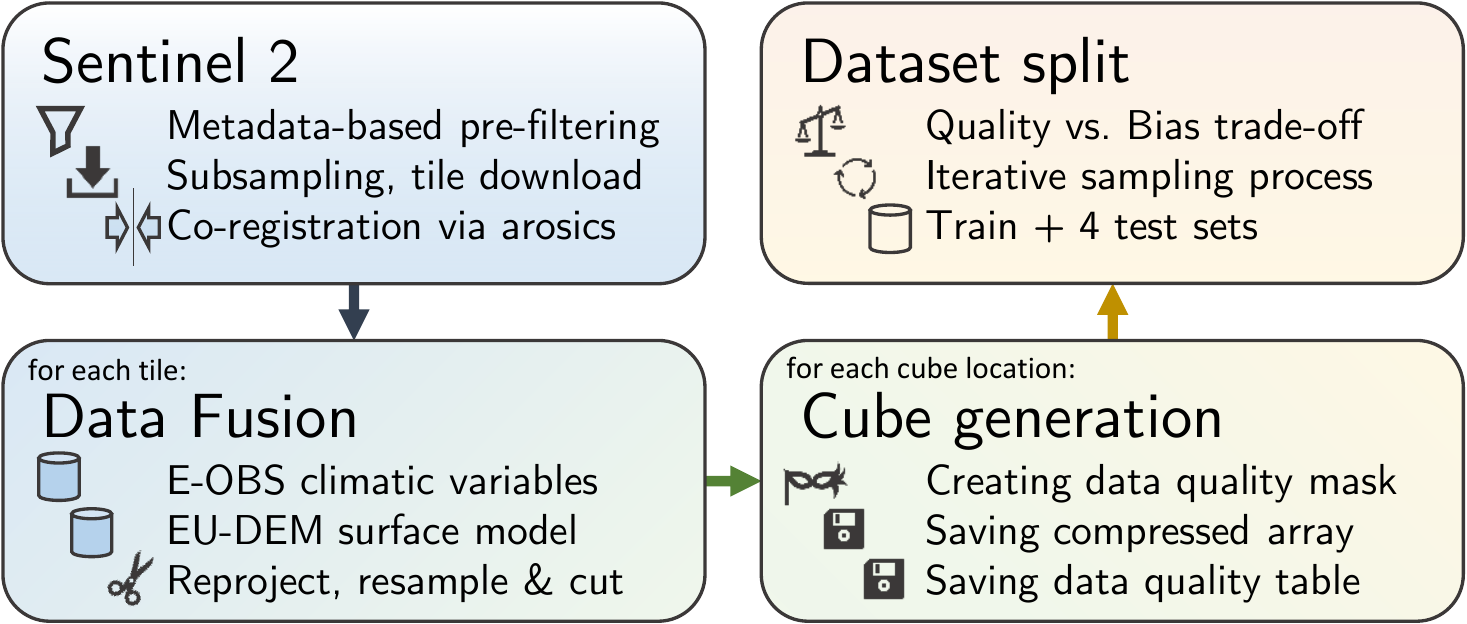}
		\caption{The dataset generation scheme of EarthNet2021.}
		\label{fig:datagen}
	\end{figure}
\end{center}

\vspace{-32pt}
\subsection{Generation}
\label{par:datagen}

\PAR{Challenges.}
In general, geospatial datasets are not analysis-ready for standard computer vision. While the former often contain large files together with information about the projection of the data, the latter requires many small data samples on an Euclidean grid. With EarthNet2021 we aim to bridge the gaps and transform geospatial data into analysis-ready samples for deep learning. To this end, we had to gather the satellite imagery, combine it with additional predictors, generate individual data samples and split these into training and test sets -- challenges which are described in the following paragraphs and lead to our dataset generation scheme, see Fig.~\ref{fig:datagen}.

\PAR{Obtaining Sentinel 2 satellite imagery.}
Downloading the full archive of Sentinel 2 imagery over Europe would require downloading Petabytes, rendering the approach unfeasible. Luckily pre-filtering is possible as the data is split by the military grid reference system into so-called tiles and for each tile metadata can be obtained from the AWS Open Data Registry\footnote{\url{https://registry.opendata.aws/sentinel-2/}} before downloading. We pre-filter and only download a random subset of 110 tiles with at least $80\%$ land visible on the least-cloudy day and minimum $90\%$ data coverage. For each tile we download blue, green, red, near-infrared and scene-classification bands over the time series corresponding to the 5-day interval with the lowest off-nadir angle. We use the sentinel-hub library\footnote{\url{https://sentinelhub-py.readthedocs.io/}} to obtain an archive of over $30$ TB raw imagery from November 2016 to May 2020. In it we notice spatial jittering between consecutive Sentinel 2 intakes, possibly due to the tandem satellites not being perfectly co-registered. We try to compensate this artifact by co-registering the time series of each tile. We use the global co-registration from the arosics library\footnote{\url{https://pypi.org/project/arosics/}} \cite{scheffler2017} inside a custom loop.

\PAR{Data fusion with E-OBS and EU-DEM.}
For each of the 110 tiles we fuse their time series with additional data. More particularly we gathered E-OBS\footnote{\url{https://surfobs.climate.copernicus.eu/}} weather variables (daily mean temperature ($TG$); daily minimum temperature ($TN$); daily maximum temperature ($TX$); daily precipitation sum ($RR$); and daily averaged sea level pressure ($PP$)) at $11.1$ km resolution and the EU-DEM\footnote{\tt\href{https://eea.europa.eu/data-and-maps/data/eu-dem/}{eea.europa.eu/data-and-maps/data/eu-dem/}} digital surface model at $25$ m resolution. We re-project, resample and cut them to two data windows. The high-resolution window has $20$ m ground resolution, matching the Sentinel 2 imagery, and the mesoscale window has $1.28$ km ground resolution. 

\PAR{Generation of minicubes.}
Given the fused data, we create a target minicube grid, cutting each tile into a regular spatial grid and a random temporal grid. Spatially, high-resolution windows of minicube do not overlap, while mesoscale windows do. Temporally, minicubes at the same location never overlap. For each location in the minicube grid, we extract the data from our fused archive, generate a data quality (i.e. cloud) mask based on heuristic rules (similar to \cite{meraner2020}) and save the minicube in a compressed numpy  array \cite{harris2020}. We also generate data quality indicators; these will be useful for selecting cubes during dataset splitting.

\begin{center}
	\begin{figure}[t]
		\centering
		\includegraphics[width=\columnwidth]{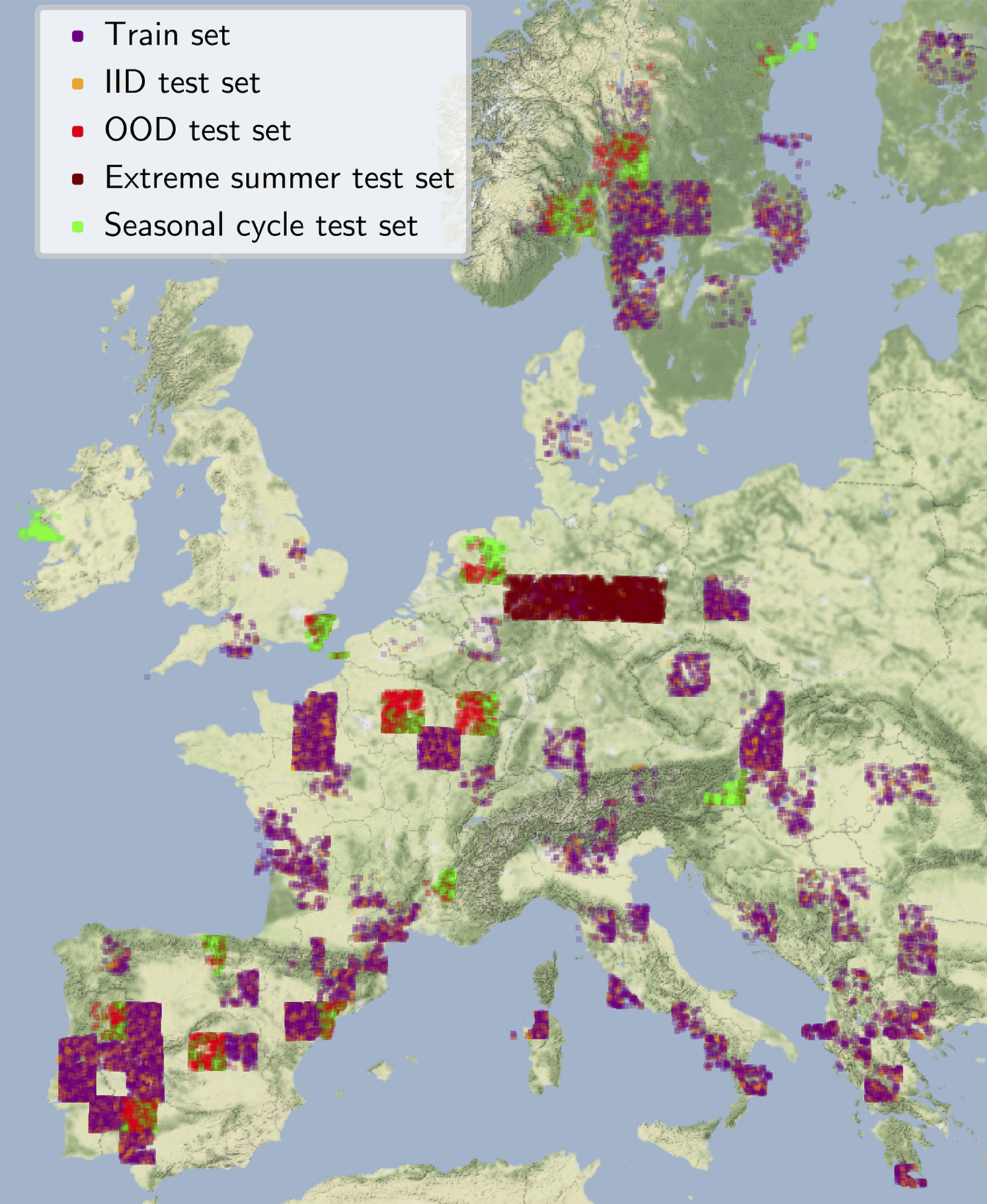}
		\caption{Spatial distribution of the samples in EarthNet2021.}
		
		\label{fig:spatial_bias}
	\end{figure} 
\end{center}

\vspace{-12pt}
\PAR{Creating the dataset split.}
In the raw EarthNet2021 corpus are over $1.3$ million minicubes. Unfortunately, most of them are of very low quality, mainly because of clouds. Now just taking minicubes above a certain quality threshold creates another problem: selection bias. To give an intuitive example: most frequently, high-quality (cloud-free) samples are found during summer on the Iberian Peninsula, whereas there barely are 4 consecutive weeks without clouds on the British Islands. We address this trade-off by introducing an iterative filtering process. Until 32000 minicubes are collected, we iteratively loosen quality restrictions for selecting high quality cubes and filling them up to obtain balance among starting months and between northern and southern geolocations. In a similar random-iterative process we separate $15$ tiles from all downloaded tiles to create a spatial out-of-domain (OOD) test set (totalling $4214$ minicubes) and randomly split the remainder tiles into $23904$ minicubes for training and $4219$ for in-domain (IID) testing.

\subsection{Description}

\begin{center}
	\begin{figure}[t]
		\centering
		\includegraphics[width=\columnwidth]{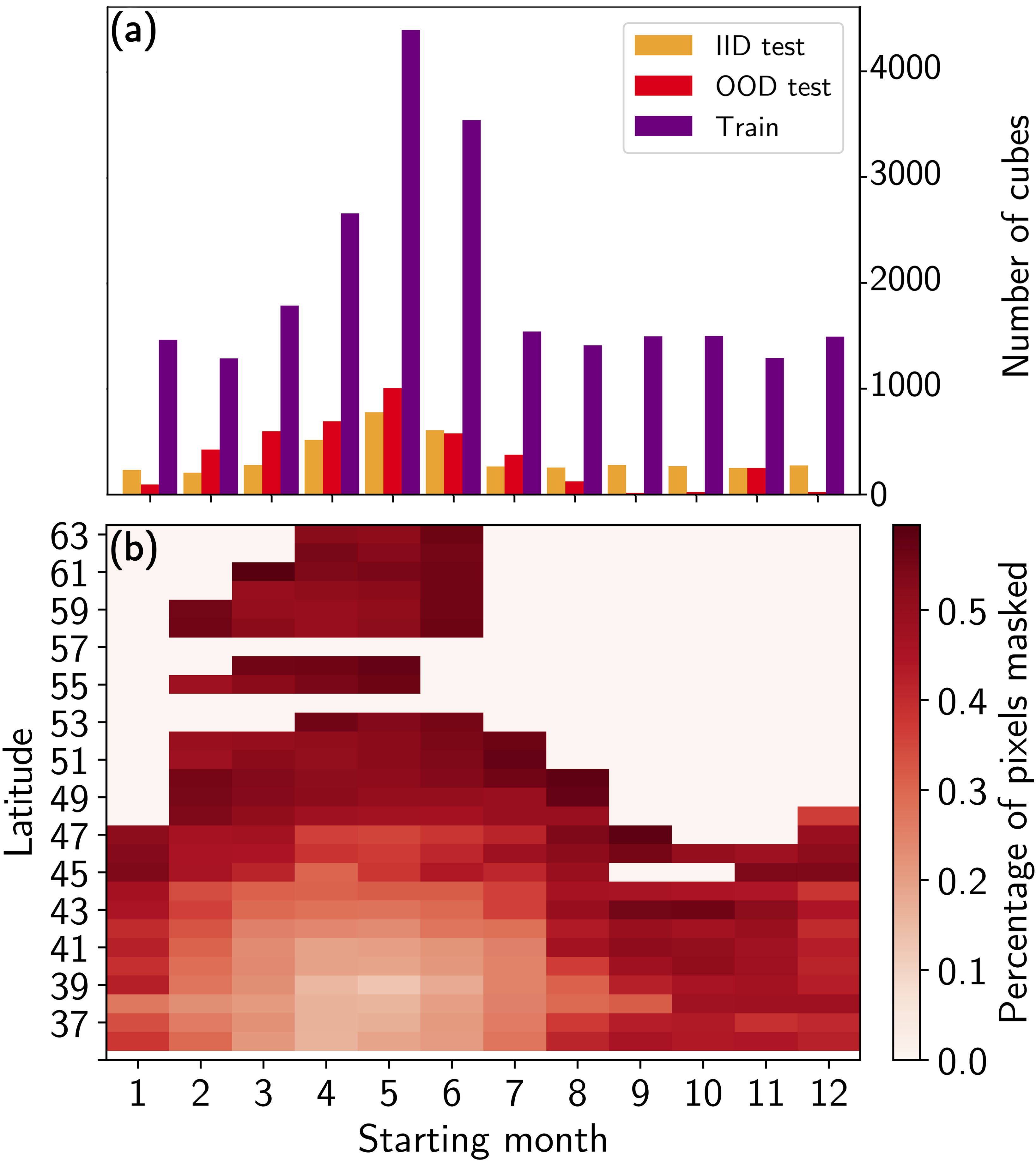}
		\caption{Monthly bias of samples. (a) Shows the monthly number of minicubes and (b) shows the data quality measured by the percentage of masked (mainly cloudy) pixels over both, months and latitude.}
		
		\label{fig:monthly_bias}
	\end{figure} 
\end{center}

\PAR{Statistics.}
The EarthNet2021 dataset spans across wider Central and Western Europe. Its training set contains $23904$ samples from $85$ regions (Sentinel 2 tiles) in the spatial extent. Fig.~\ref{fig:spatial_bias} visualizes the spatial distribution of samples. $71\%$ of the minicubes in the training set lay in the southern half, which also contains more landmass. In the northern half we observe a strong clustering in the vicinity of the Oslofjord, which is possibly random. Temporally most minicubes cover the period of May to October (Fig.~\ref{fig:monthly_bias}a). While this certainly biases the dataset, it might actually be desirable because some of the most devastating climate impacts (e.g., heatwaves, droughts, wildfires) occur during summer. Fig.~\ref{fig:monthly_bias}b shows the bias-quality trade-off, observe that most high quality minicubes are from summer in the Mediterranean. Also, it shows that EarthNet2021 does not contain samples covering winter in the northern latitudes. This is possibly an effect of our very restrictive quality masking wrongly classifying snow as clouds.

\PAR{Comparison to other datasets.}
Earth surface forecasting is a novel task, thus there are no such datasets prior to EarthNet2021. Still, since it also belongs to the broader set of analysis-ready datasets for deep learning, we can assert that it is large enough for training deep neural networks. In supplementary table~2
we compare a range of datasets using either satellite imagery or being targeted to video prediction models. By pure sample size, EarthNet2021 ranks solid, yet, the number is misleading since individual samples are different. By additionally comparing the  size in gigabytes, we assert that EarthNet2021 is indeed a large dataset.

\PAR{Limitations.}
Clearly, EarthNet2021 limits models to work solely on Earth surface forecasting in Europe. Additionally, the dataset is subject to a selection bias; therefore, there are areas in Europe for which model generalizability could be problematic. Furthermore, EarthNet2021 leverages observational products instead of actual forecasts. Thus, while this certainly is practical for a number of reasons, Earth surface models trained on EarthNet2021 should be viewed as experimental and might not be plug-and-play into production with seasonal weather model forecasts.

\section{EarthNet2021 Challenge}\label{sec:challenge}
\subsection{Overview}
\begin{center}
	\begin{figure}[t]
		\centering
		\includegraphics[width=\columnwidth]{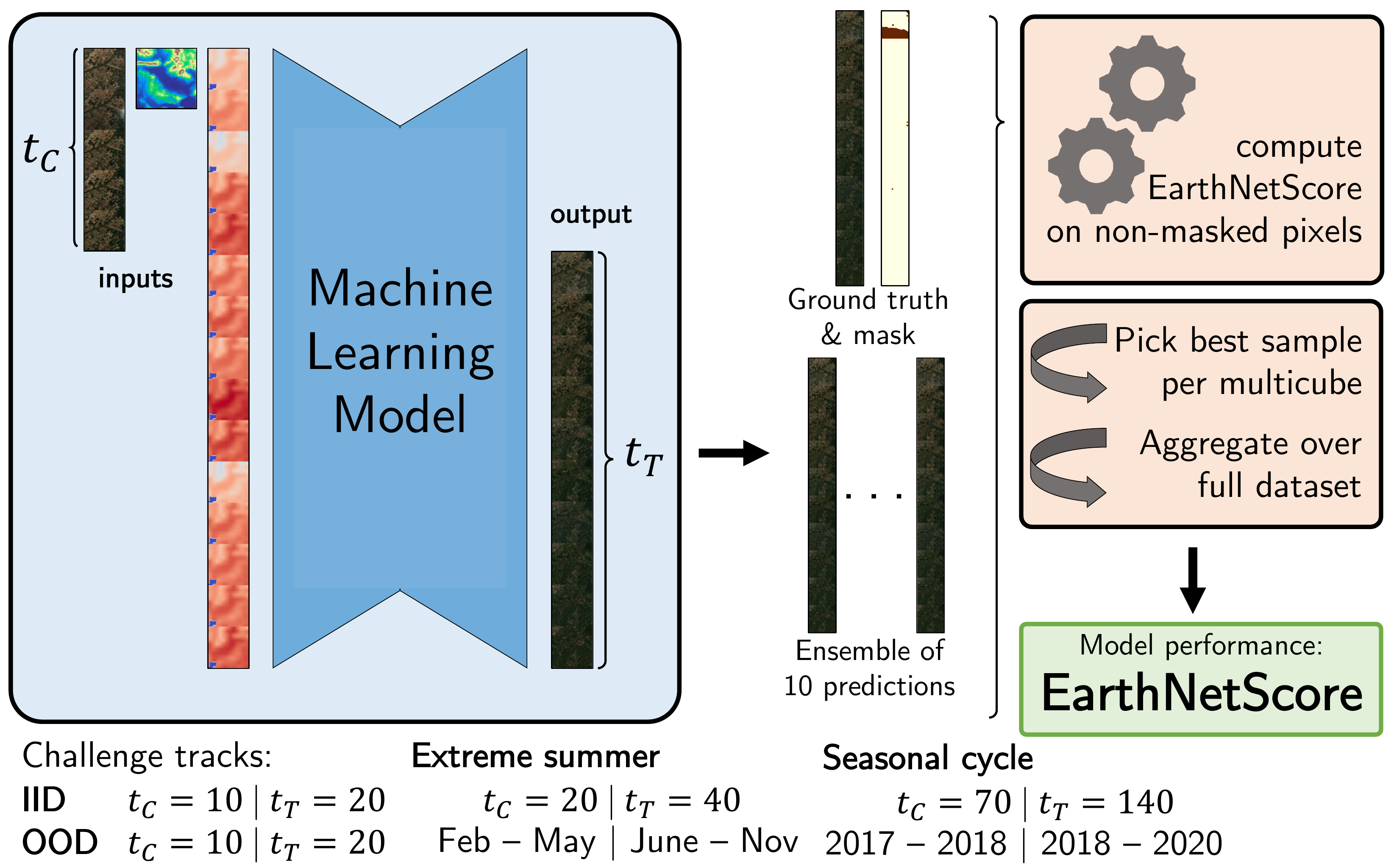}
		\caption{Evaluation pipeline for models on EarthNet2021. For predicting the $t_T$ target frames, a model can use satellite images from the $t_C$ context frames, the static DEM and mesoscale climatic variables including those from the target time steps.}
		\label{fig:eval}
	\end{figure}
\end{center}

\PAR{Model intercomparison.}
Modeling efforts are most useful when different models can easily be compared. Then, strengths and weaknesses of different approaches can be identified and state-of-the-art methods selected. We propose the EarthNet2021 challenge as a model intercomparison exercise in Earth surface forecasting built on top of the EarthNet2021 dataset. This motivation is reflected in the design of the challenge. We define an evaluation protocol by which approaches can be compared as well as provide a framework, such that knowledge between modelers is easily exchanged. There is no reward other than scientific contribution and a publicly visible leaderboard. Evaluating Earth surface forecasts is not trivial. Since it is a new task, there is not yet a commonly used criterion. We design the EarthNetScore as a ranking criterion balancing multiple goals and center the evaluation pipeline around it (see Fig.~\ref{fig:eval}). Moreover, we motivate four challenge tracks. These allow comparison of models' validity and robustness and applicability to extreme events and the vegetation cycle.

\PAR{EarthNet2021 framework.}
To facilitate research we provide a flexible framework which kick-starts modelers and removes double work between groups. The evaluation pipeline is packaged in the EarthNet2021 toolkit and leverages multiprocessing for fast inference. Additionally, challenge participants are encouraged to use the model intercomparison suite. It shall give one entry point for running a wide range of models. Currently it features model templates in PyTorch and TensorFlow and additional graphical output useful for debugging and visual comparison.

\subsection{EarthNetScore}
\PAR{Components.}
Evaluating Earth surface predictions is non-trivial. Firstly, because the topological space of spectral images is not a metric space, secondly, because clouds and other data degradation hinder evaluation, and, thirdly, because ranking simply by root mean squared error might lead to ordinal rankings of imperfect predictions that experts would not agree to. Instead of using a single score, we define the \emph{EarthNetScore} $ENS$ by combining multiple components. As the first component, we use the median absolute deviation $MAD$. It is a robust distance in pixel-space which is justified by the goal that predicted and target values should be close. Secondly, $OLS$, the difference of ordinary least squares linear regression slopes of pixelwise Normalized Difference Vegetation Index (NDVI) timeseries, gives an indicator as to whether the predictions are able to reproduce the trend in vegetation change. This together with the Earth mover distance $EMD$ between pixelwise NDVI time series over the short span of 20 time steps is a good proxy of the fit (in the distribution and direction) of the vegetation time series. The time series based metrics $OLS$ and $EMD$ are also largely robust to missing data points, which poses a consistency constraint on model predictions at output pixels for which no target data is available. Finally, the structural similarity index $SSIM$ is a perceptual metric imposing predicted frames to have similar spatial structure to the target satellite images. All component scores are modified to work properly in the presence of a data quality mask, rescaled to match difficulty and transformed to lay between 0 (worst) and 1 (best).

\PAR{Computation.}
Combining these components is another challenge. We would like to define the EarthNetScore as:
\begin{align}\label{eq:ens}
ENS = \frac{4}{(\frac{1}{MAD} + \frac{1}{OLS} + \frac{1}{EMD} + \frac{1}{SSIM})}.
\end{align}
This is the harmonic mean of the four components, so it lays strongly to the worst performing component. Yet, computing $ENS$ over a full test set requires further clarification. Earth surface forecasting is a stochastic prediction task, models are allowed to output an ensemble of predictions. Thus, for each minicube in the test set there might multiple predictions (up to 10). In line with what is commonly done in video prediction, we compute the subscores for each one of them but, only take the prediction for which equation~\ref{eq:ens} is highest for model intercomparison. In other words, the evaluation pipeline only accounts for the best prediction per minicube. This is superior to average predictions as it allows for an ensemble of discrete, sharp, plausible outcomes, something desired for Earth surface forecasting given its highly multimodal nature. Still, this evaluation scheme, suffers severely from not being able to rank models according to their modeled distribution. Once the components for the best predictions for all minicubes in the dataset are collected, we average each component and then calculate the $ENS$ by feeding the averages to equation~\ref{eq:ens}. Then, the $ENS$ ranges from $0$ to $1$, where $1$ is a perfect prediction.

\begin{table*}[t]
\setlength{\tabcolsep}{5pt}
\centering
\begin{tabularx}{\textwidth}{Xccccccccccccc}
\toprule
                   & \multicolumn{6}{c}{\textbf{IID}}                                                                                                                                               & \hspace{0.5cm} & \multicolumn{6}{c}{\textbf{OOD}}                                                                                                               \\ 
 & \textbf{ENS} & \quad & \textbf{MAD} & \textbf{OLS} & \textbf{EMD} & \textbf{SSIM} & $\mkern1mu$ & \textbf{ENS}              & $\mkern1mu$ & \textbf{MAD}             & \textbf{OLS}               & \textbf{EMD}               & \textbf{SSIM}              \\ 
 \midrule
Persistence       & 0.2625                           & & 0.2315                           & 0.3239                           & 0.2099                           & 0.3265     & & 0.2587 && 0.2248 & 0.3236 & 0.2123 & 0.3112 \\
Channel-U-Net      & 0.2902                           & & 0.2482                           & 0.3381                           & 0.2336                           & 0.3973         & & 0.2854                     & & 0.2402                     & 0.3390                      & 0.2371                     & 0.3721                     \\
Arcon          & 0.2803                           & & 0.2414                            & 0.3216                           & 0.2258                           & 0.3863        & & 0.2655                     & & 0.2314                     & 0.3088                     & 0.2177                     & 0.3432     \\   
\bottomrule
\end{tabularx}
\vspace*{4pt}
\caption{Models performance on EarthNet2021. See supplementary material for Extreme and Seasonal test sets.}
\label{tab:baseline}
\end{table*}

\subsection{Tracks}
\PAR{Main (IID) track.}
The EarthNet2021 main track checks model validity. It uses the IID test set, which has minicubes that are very similar (yet randomly split) as those seen during training. Models get 10 context frames of high resolution 5-daily multispectral satellite imagery (time [t-45, t]), static topography at both mesoscale and high resolution, and mesoscale dynamic climate conditions for 150 past and future days (time [t-50, t+100]). Models shall output 20 frames of high-resolution sentinel 2 bands red, green, blue and near-infrared for the next 100 days (time [t+5,t+100]). These predictions are then evaluated with the EarthNetScore on cloud-free pixels from the ground truth. This track follows the assumption that, in production, any Earth surface forecasting model would have access to all prior Earth observation data, thus the test set has the same underlying distribution as the training set.

\PAR{Robustness (OOD) track.}
In addition to the main track, we offer a robustness track. Even on the same satellite data, deep learning models might generalize poorly across geolocations \cite{benson2020}, thus it is important to check model performance on an out-of-domain (OOD) test set. This track has a weak OOD setting; in which minicubes solely are from different Sentinel 2 tiles than those seen during training, which is possibly only a light domain shift. Still, it is useful as a first benchmark to check applicability of models outside the training domain.

\PAR{Extreme summer track.}
Furthermore, EarthNet2021 contains two tracks particularly focused on Earth system science hot topics, which should both be understood as more experimental. The extreme summer track contains cubes from the extreme summer 2018 in northern Germany \cite{bastos2020}, with 4 months of context (20 frames) starting from February and 6 months (40 frames) starting from June to evaluate predictions. For these locations, only cubes before 2018 are in the training set. Being able to accurately downscale the prediction of an extreme heat event and to predict the vegetation response at a local scale would greatly benefit research on resilience strategies. In addition, the extreme summer track can in some sense be understood as a temporal OOD setting.

\PAR{Seasonal cycle track.}
While not the focus of EarthNet2021, models are likely able to generate predictions for longer horizons. Thus, we include the seasonal cycle track covering multiple years of observations; hence, including vegetation full seasonal cycle. This track is also in line with the recently rising interest in seasonal forecasts within physical climate models. It contains minicubes from the spatial OOD setting also used for the robustness tracks, but this time each minicube comes with 1 year (70 frames) of context frames and 2 years (140 frames) to evaluate predictions. For this longer prediction length, we change the EarthNetScore $OLS$ component to be calculated over disjoint windows of 20 frames each.

\section{Models}
As first baselines in the EarthNet2021 model intercomparison, we provide three models. One is a naive averaging persistence baseline while the other two are deep learning models slightly modified for guided video prediction. Performance is reported in \tabref{tab:baseline}.

\PAR{Persistence baseline}
The EarthNet2021 unified toolkit comes with a pre-implemented baseline in NumPy. It simply averages cloud-free pixels over the context frames and uses that value as a constant prediction. Performance is shown in table \ref{tab:baseline}.

\PAR{Autorregressive Conditional video prediction baseline}
The Autorregressive Conditional  video prediction baseline (Arcon) is based on Stochastic adversarial video prediction (SAVP, \cite{lee2018stochastic}) that was originally was used as an unguided or weakly guided video prediction model. We extend SAVP for EarthNet2021 by stacking the guiding variables as extra video channels. To this end, climatic variables had to be resampled to match imagery resolution. In addition, SAVP cannot make use of the different temporal resolution of predictors and targets (daily vs. 5 daily) so predictors were reduced by taking the 5-daily mean, these steps resulted in guiding information loss. Since there is no moving objects in satellite imagery, but just a widely variable background, all SAVP components specifically designed for motion prediction were disabled. Image generation from scratch and reuse of context frames as background was enabled. Different to traditional video input data, EarthNet2021 input satellite imagery is defective, as a model shall not forecast clouds and other artifacts. Thus, different to the original implementation, we train Arcon just with mean absolute error over non-masked pixels; in particular, this means no adversarial loss was used. 

\begin{figure}[t]
	\centering
	\includegraphics[width=\columnwidth]{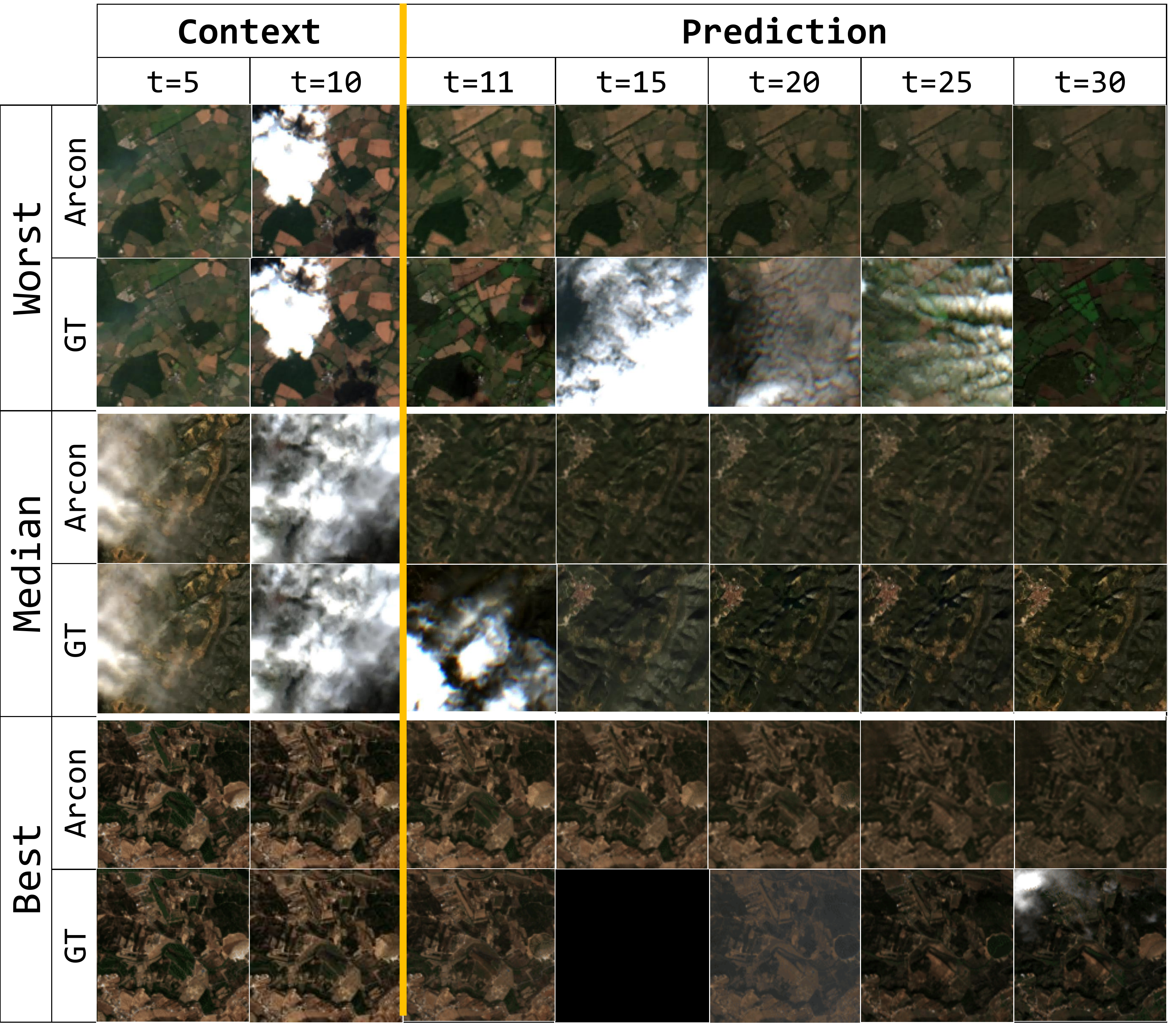}
	\caption{Worst, median and best samples predicted by Arcon according to EarthNetScore over the full IID test set. Yellow line marks the beginning of the predicted frames.}
	\label{fig:savp}
\end{figure}

Arcon outperforms the persistence baseline in every test set except the full seasonal cycle test (see table \ref{tab:baseline} for IID and OOD results), where, possibly the model breaks down when fed a context length higher than 10. The model shows degrading forecasting performance at the longer temporal horizon (see Fig. \ref{fig:savp}). These results give us two hints. First, it is necessary to overhaul and adapt current video prediction models to make them capable of tackling the strongly guided setting. Second, since the slightly adapted SAVP shows skill over the  persistence baseline, we can anticipate current video prediction models to be a useful starting point.

\PAR{Channel-U-Net baseline}

This architecture is inspired by the winning solution to the 2020 Traffic4cast challenge \cite{choi2020utilizing}. Traffic map forecasting is to a certain degree similar to the proposed task of Earth surface forecasting. The solution used a U-Net architecture \cite{ronneberger2015u} with dense connections between layers. All available context time step inputs were stacked along the channel dimension and fed into the network. Subsequently the model outputs all future time steps stacked along the channel dimension, which are then reshaped for proper evaluation. We call such an approach a Channel-U-Net. Here we present a Channel-U-Net with an ImageNet \cite{deng2009imagenet} pre-trained DenseNet161 \cite{iandola2014densenet} encoder from the Segmentation Models PyTorch library\footnote{\url{https://smp.readthedocs.io/}}. As inputs we feed the center 2x2 meteorological predictors upsampled to full spatial resolution, the high-resolution DEM and all the satellite channels from the 10 context time steps, resulting in 191 input channels. The model outputs 80 channels activated with a sigmoid, corresponding to the four color channels for each of the 20 target time steps. We trained the model for 100 Epochs on a quality masked L1 loss with Adam \cite{kingma2014adam}, an initial learning rate of 0.002, decreased by a factor 10 after 40, 70 and 90 epochs. We use a batch size of 64 and 4 x V100 16GB GPUs. For the extreme and the seasonal tracks we slide the model through the time series by feeding back its previous outputs as new inputs after the initial prediction, which uses the last 10 frames of context available.

Channel-U-Net is the overall best performing model, even though it does not model temporal dependencies explicitly. This model also underperforms the persistence baseline on the seasonal test set, possibly due to the sliding window approach used for the much longer prediction length.

\section{Outlook}
If solved, Earth surface forecasting will greatly benefit society by providing seasonal predictions of climate impacts at a local scale. These are extremely informative for implementing preventive mitigation strategies. EarthNet2021 is a stepping-stone towards the collaboration that is necessary between modelers from Computer Vision and domain experts from the Earth System Sciences. The dataset requires developing guided video prediction models, a unique opportunity for video prediction researchers to extend their approaches. Since the guided setting allows modeling in a more controlled environment, there is the possibility, that gained knowledge can also be transferred back to general (unguided) video prediction. Eventually, numerical Earth System Models \cite{eyring2016overview} could benefit from the data-driven high-resolution modelling by Earth surface forecasting models. In so-called hybrid models \cite{reichstein2019}, both components could be combined.

EarthNet2021 is the first dataset for spatio-temporal Earth surface forecasting. As such, it comes with a number of limitations, including some that will be discovered during model development. Through the EarthNet2021 framework, especially the model intercomparison suite, we hope to create a space for communication between different stakeholders. Then, we could remove pressing issues iteratively.  We hope EarthNet2021 will serve as a starting point for community building around high-resolution Earth surface forecasting.

{
\vspace{12pt}
\small
\noindent
\textbf{Acknowledgements}
We thank three anonymous reviewers and the area chair for their constructive reviews. We are grateful to the Centre for Information Services and High Performance Computing [Zentrum für Informationsdienste und Hochleistungsrechnen (ZIH)] TU Dresden for providing its facilities for high throughput calculations.

\noindent
\textbf{Authors contribution}
\textbf{C.R.} Experimental design, EarthNet model intercomparison suite, baseline persistence and Arcon models, website, documentation, manuscript.
\textbf{V.B.} Experimental design, dataset creation pipeline, EarthNet toolkit package, baseline channel-net model, website, documentation, manuscript.
\textbf{M.R.} Experimental design, metric selection,  manuscript revision.
\textbf{J.R.} Experimental design, manuscript revision.
\textbf{J.D.} Experimental design, metric selection, manuscript revision.

\noindent
\textbf{Code and data availability}
All information is provided on our website \url{www.earthnet.tech}. 
}

\newpage
{\small
\bibliographystyle{ieee_fullname}
\bibliography{refs}
}


\label{par:suplementary}

\section{Supplementary material}

\begin{figure*}[h!]
	\centering
	\includegraphics[width=\columnwidth]{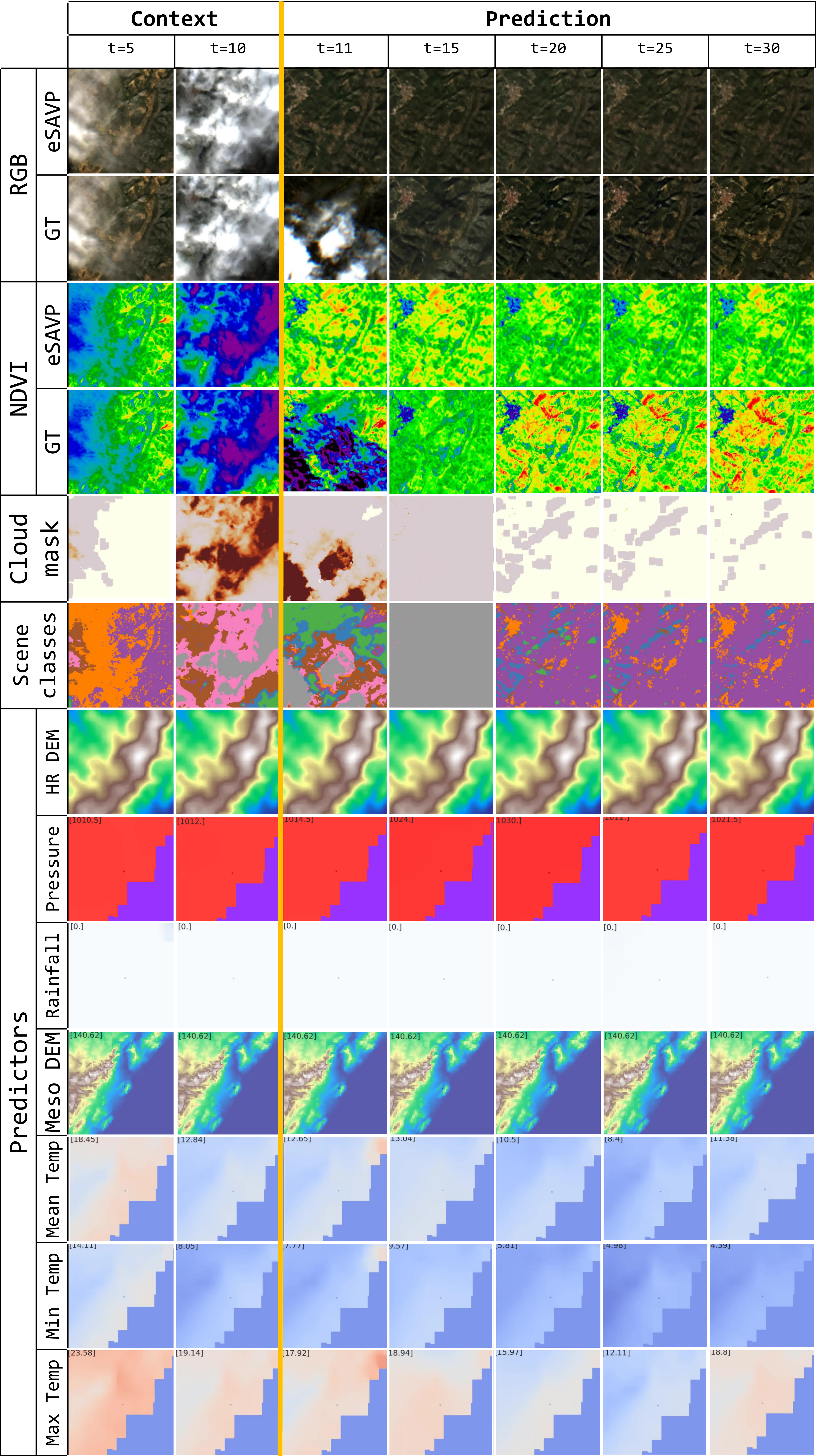}
	\caption{Median sample predicted by the extended SAVP model (Arcon) according to EarthNetScore over the full IID set. Predictions of Arcon on "eSAVP" rows, ground truth on "GT" rows. Yellow line marks the beginning of the predicted frames. Predictors and auxiliary layers are plotted directly from the minicube file.}
	
	\label{fig:savp_guided}
\end{figure*}

\PAR{Arcon visualizations}
Median scoring sample according to ENS over the IID test set with all guiding variables can be seen in Fig.~\ref{fig:savp_guided}.

\PAR{Extra baseline results}
Extreme and Seasonal test set results for the tested models can be seen in table \ref{tab:baseline2}.

\PAR{Similar datasets}
Table \ref{tab:datasets} contains information on datasets similar to EarthNet2021.

\PAR{Dataset information}
Table \ref{tab:variables} contains information about the variables used in EarthNet2021. Table \ref{tab:npz} contains information about the packaging of minicubes into files.

{
	\begin{table}[bh!]
		\footnotesize
		\addtolength{\tabcolsep}{-0.15em}
		\centering
		\begin{tabularx}{\columnwidth}{Xlcc}
			\toprule
			Dataset & Task & Samples & Size(GB)  \\
			\midrule
			\multicolumn{4}{l}{$\qquad$ \emph{Satellite imagery}}\\
			SpaceNet7 \cite{van2018spacenet} & Building footprint det. & 2.4k & 56 \\
			AgriVision \cite{chiu2020agriculture} & Field anomaly det. & 21k & 4 \\
			xBD \cite{gupta2019creating} & Building damage det. & 22k & 51 \\
			
			DynamicEarthNet\cite{zhu_qiu_hu_camero} & Land cover change det. & 75*300 & 63 \\
			
			BigEarthNet \cite{sumbul2019bigearthnet} & Land use classification & 590k & 121 \\
			\midrule
			\multicolumn{4}{l}{$\qquad$ \emph{Video prediction}}\\
			Cityscapes \cite{cordts2016cityscapes} & Video annotation & 25k & 56 \\
			Traffic4cast \cite{kreil2020} & Traffic forecasting & 474 & 23 \\
			UCF101 \cite{soomro2012ucf101} & Human actions &  13k & 7 \\
			\midrule
			EarthNet2021 & Earth surface forecasting & 32k*30 & 218 \\
			
			\bottomrule
		\end{tabularx}
		\vspace*{4pt}
		\caption{Large-scale deep-learning-ready datasets.}
		\label{tab:datasets}
	\end{table}
}

\begin{table*}[h!]
\setlength{\tabcolsep}{5pt}
\centering
\begin{tabularx}{\textwidth}{Xccccccccccccc}
\toprule
                   & \multicolumn{6}{c}{\textbf{Extreme}}                                                                                                                                               & \hspace{0.5cm} & \multicolumn{6}{c}{\textbf{Seasonal}}                                                                                                               \\ 
 & \textbf{ENS} & \quad & \textbf{MAD} & \textbf{OLS} & \textbf{EMD} & \textbf{SSIM} & $\mkern1mu$ & \textbf{ENS}              & $\mkern1mu$ & \textbf{MAD}             & \textbf{OLS}               & \textbf{EMD}               & \textbf{SSIM}              \\ 
 \midrule
Persistence       & 0.1939                           & & 0.2158                           & 0.2806                           & 0.1614                           & 0.1605     & & 0.2676 && 0.2329 & 0.3848 & 0.2034 & 0.3184 \\
Channel-U-Net      & 0.2364                           & & 0.2286                           & 0.2973                           & 0.2065                           & 0.2306         & & 0.1955                     & & 0.2169                     & 0.3811                      & 0.1903                     & 0.1255                     \\
Arcon          & 0.2215                           & & 0.2243                            & 0.2753                           & 0.1975                           & 0.2084        & & 0.1587                     & & 0.2014                     & 0.3788                     & 0.1787                     & 0.0834     \\   
\bottomrule
\end{tabularx}
\vspace*{4pt}
\caption{Models performance on EarthNet2021 over the extreme and seasonal test sets}
\label{tab:baseline2}
\end{table*}

\begin{table*}[h!]
	\begin{tabularx}{\textwidth}{Xcccc}
		\toprule
		Variable & Description & Source & Rescaling & Unit  \\ 
		\midrule
		\textcolor{blue}{b} & 490nm reflectance & Sentinel-2 MSI L2A\cite{louis2016} & $None$ & TOA [0,1]\\
		\textcolor{green}{g} & 560nm reflectance & Sentinel-2 MSI L2A\cite{louis2016} & $None$ & TOA [0,1]\\
		\textcolor{red}{r} & 665nm reflectance & Sentinel-2 MSI L2A\cite{louis2016} & $None$ & TOA [0,1]\\
		\textcolor{magenta}{nif} & 842nm reflectance & Sentinel-2 MSI L2A\cite{louis2016} & $None$ & TOA [0,1]\\
		cld & Cloud probability & Sentinel-2 Product\cite{louis2016} & $None$ & \%\\
		scl & Scene classification & Sentinel-2 Product\cite{louis2016} & $None$ & categorical\\
		cldmask & Cloud mask & EarthNet pipeline & $None$ & binary\\
		elevation & Digital elevation model & EU-DEM\cite{bashfield2011} & $2000\cdot(2\cdot elevation-1)$ & meters\\
		rr & Rainfall & E-OBS\cite{cornes2018} & $50 \cdot rr$ & mm/d\\
		pp & Pressure & E-OBS\cite{cornes2018} & $900+200 \cdot pp$ & mbar\\
		tg & Mean temperature & E-OBS\cite{cornes2018} & $50\cdot(2\cdot tg-1)$ & celcius\\
		tn & Minimum temperature & E-OBS\cite{cornes2018} & $50\cdot(2\cdot tg-1)$ & celcius\\
		tm & Maximum temperature & E-OBS\cite{cornes2018} & $50\cdot(2\cdot tg-1)$ & celcius\\
		\bottomrule
	\end{tabularx}
	\vspace*{4pt}
	\caption{Variables included in EarthNet2021.}
	\label{tab:variables}
\end{table*}

\begin{table*}[h!]
	\begin{tabularx}{\textwidth}{Xcccc}
		\toprule
		Array name & Shape & Spatial (res) & Temporal (res) & Variables  \\ 
		\midrule
		highresdynamic &[128,128,7,$t_{hr}$] & [128,128] (20m) & (30,60,210) (5-daily) & [\textcolor{blue}{b},\textcolor{green}{g},\textcolor{red}{r},\textcolor{magenta}{nif},cld,scl,cldmsk] \\
		mesodynamic &[80,80,5,$t_{meso}$] & [80,80] (1.28km) & (150,300,150) (daily) & [rr,pp,tg,tn,tx] \\
		highresstatic &[128,128] & [128,128] (20m) & [1] (fixed) & elevation \\
		mesostatic &[80,80] & [80,80] (1.28km) & [1] (fixed) & elevation \\
		\bottomrule
	\end{tabularx}
	\vspace*{4pt}
	\caption{Content on each compressed 'sample.npz' minicube files.}
	\label{tab:npz}
\end{table*}

\end{document}